\documentclass[11pt]{article}
\usepackage{konvens2016}
\usepackage{mathptmx}
\usepackage{booktabs}
\usepackage[scaled=.90]{helvet}
\usepackage{courier}
\usepackage{graphicx}
\usepackage{url}
\usepackage{latexsym}
\usepackage{multirow}
\usepackage{booktabs}
\usepackage{color}
\usepackage{url}

\title{What to do about non-standard (or \textit{non-canonical}) language in NLP}

\author{Barbara Plank \\
  University of Groningen \\
  {\tt b.plank@rug.nl}}

\date{}

\begin{document}
\maketitle
\begin{abstract}
Real world data differs radically from the benchmark corpora we use in natural language processing (NLP).
As soon as we apply our technologies to the real world, performance drops. The reason for this problem is obvious: NLP models are trained on samples from a limited set of \textit{canonical} \textit{varieties} that are considered \textit{standard}, most prominently English newswire. 
However, there are many dimensions, e.g., socio-demographics, language, genre, sentence type, etc.\ on which texts can differ from the standard.
The solution is not obvious: we cannot control for all factors, and it is not clear 
how to best go beyond the current practice of training on homogeneous data from a single domain and language.

In this paper, I review the notion of canonicity, and how it shapes our community's approach to language.
I argue for leveraging what I call \textit{fortuitous data}, i.e., non-obvious data that is 
hitherto neglected, hidden in plain sight, or raw data that needs to be refined. 
If we embrace the variety of this heterogeneous data by combining it with proper algorithms, we will not only produce more robust models, but will also enable adaptive language technology capable of addressing natural language variation. 
\end{abstract}

\section{Introduction}
The publication of the Penn Treebank Wall Street Journal (WSJ) corpus in the late 80s has undoubtedly pushed NLP from symbolic computation to statistical approaches, which dominate our field up to this day. The WSJ has become the NLP benchmark dataset for many tasks (e.g., part-of-speech tagging, parsing, semantic role labeling, discourse parsing), and has developed into the \textit{de-facto} ``standard'' in our field. 

However, while it has advanced the field in so many ways, it has also introduced almost imperceptible biases: why is newswire considered more standard or more canonical than other text types? Journalists are trained writers who make fewer errors and adhere to a codified norm.\footnote{We do not explicitly concern us here with issues of language prescription, but rather on the assumption-heavy perceptions of some instances of language as `more normal'. }  But let us pause for a minute. 
If NLP had emerged only in the last decade, 
would newswire data still be our canon? Or would, say, Wikipedia be considered canonical? User-generated data is less standardized, but is highly available. 
If we take this thought further and start over today, maybe we would be in an `inverted' world:  social media is standard and newswire with its `headlinese' is the `bad language'~\cite{eisenstein:2013:bad}. It is easy to collect large quantities of social media data. Whatever we consider canonical, all data comes with its biases, even more democratic media like Wikipedia carry their own peculiarities.\footnote{For instance, the demographics of Wikipedia shows that mostly young single men aged 18-30 contribute, see \url{https://strategy.wikimedia.org/wiki/Wikimedia_users#Demographics}} 

\begin{figure}
\includegraphics[width=0.49\textwidth]{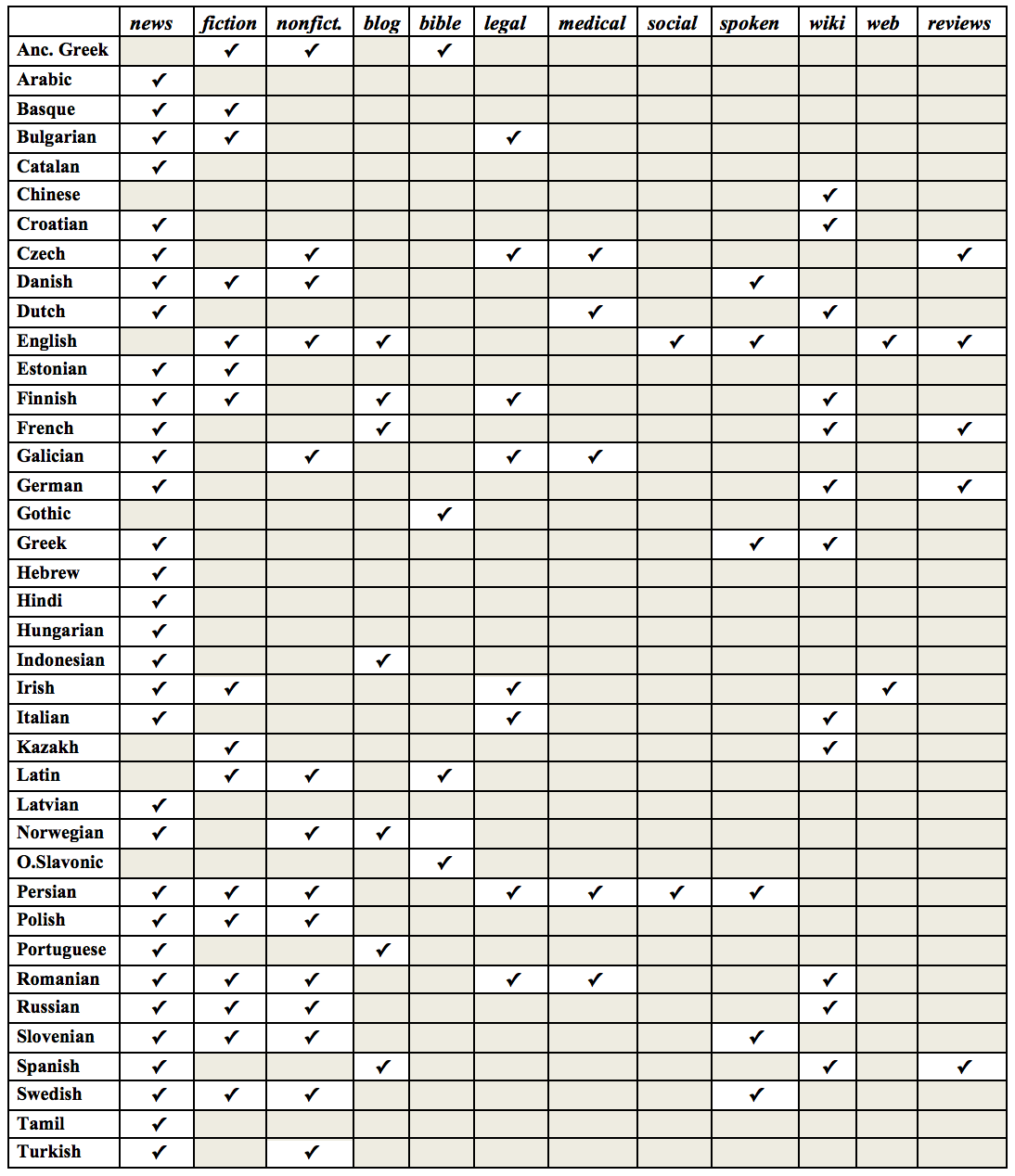}
\caption{The problem of \textit{training data sparsity} illustrated for parsing: available annotated data in languages and domains; subset of syntactically-annotated treebanks from Universal Dependencies v1.3 for which domain/genre info was available. } %Note the news column. For UDv1.3 all languages for which metadata is available contain newswire, except for one (Khazak, Wikipedia).}
\label{fig:domains}
\end{figure}

It seems that what is considered canonical hitherto is mostly a historical coincidence and motivated largely by availability of resources. Newswire has and actually still \textit{does} dominate our field. For example, in Figure~\ref{fig:domains}, I plot domains versus languages for the treebank data in version 1.3 of the on-going Universal Dependencies\footnote{\url{http://universaldependencies.org/}} project~\cite{universal1.2}. Almost all languages include newswire, except ancient languages (for obvious reasons), English (since most data comes from the Web Treebank) and Khazak, Chinese (Wikipedia).  
While including other domains and languages is highly desirable, it is impossible to find unbiased data.\footnote{This is related to the problem of overexposure in ethics, e.g.,~\cite{hovy:spruit:2016}.} Let's be \textit{aware} of this fact and try to collect enough biased data. 

\textit{Processing} non-canonical (or non-canonical) data is difficult.
A series of papers document large drops in accuracy when moving across domains~\cite[inter alia]{mcclosky:2010,foster:ea:2011}. There is a large body of work focusing on correcting for domain differences. Typically, in domain adaptation (DA) the task is to adapt a model trained on some source domain to perform better on some new target domain. However, it is less clear what really folds into a \textit{domain}. In Section~\ref{sec:variationspace}, I will review the notion of domain and propose what I call \textit{variety space}.  

Is the \textit{annotation} of non-canonical also more difficult, just like its processing appears to be? Processing and annotating are two aspects, and 
the difficulty in one, say processing, does not necessarily propagate the same way to annotation~\cite{plank-martinezalonso-sogaard:2015:LAW}. 
However, very little work exists on disentangling the two. The same is true for examining what really constitutes a domain. What remains is clear: the challenge is all about  \textit{variations} of data. Language continuously changes, for various reasons (different social groups, communicative purposes, changes over time), and so we will continuously face interesting challenges, both for processing and annotation. 

In the remainder I will look at the NLP community's approach to face these challenges. 
I will outline one potential way to go about it, arguing for the use of \textit{fortuitous data}, and end by returning to the question of  domain.

%\begin{figure}
%\resizebox{\columnwidth}{!}{
%\begin{tabular}{lccccccccccc}
%\toprule
%              & news & fiction & nonfiction & blog & medical & legal & social & spoken & wiki & web & reviews \\
%              \midrule
%Basque   & $\bullet$ & $\bullet$ \\ 
%Bulgarian & $\bullet$ & $\bullet$ & & & & $\bullet$\\ 
%Croatian & $\bullet$\\ 
%Czech   & $\bullet$\\ 
%Danish & $\bullet$\\ 
%Dutch & $\bullet$\\ 
%English & $\bullet$\\ 
%Finnish & $\bullet$\\ 
%French & $\bullet$\\ 
%German & $\bullet$\\ 
%Greek & $\bullet$\\ 
%Hungarian & $\bullet$\\ 
%Irish & $\bullet$\\ 
%Italian & $\bullet$\\ 
%Spanish & $\bullet$\\ 
%Swedish & $\bullet$\\ 
%\bottomrule
%\end{tabular}
%}
%\end{figure}
%%
%%
\section{What to do about non-standard data}
There are generally three main approaches to go about non-standard data.

\subsection{Annotate more data}\label{sec:annotation}
Annotating more data is a first and intuitive solution. However, it is na\"{i}ve, for several reasons. 

\textit{Domain} (whatever that means) and \textit{language} (whatever that comprises) are two factors of text variation. 
Now take the cross-product between the two. We will never be able to create annotated data that spans all possible combinations. This is the problem of \textit{training data sparsity},  illustrated in Figure~\ref{fig:domains}. The figure only shows a tiny subset of the world's languages, and a tiny fraction of potential \textit{domains} out there. The problem is that most of the data that is available out there is unlabeled. Annotation requires time. At the same time, ways of communication change, so what we annotate today might be very distant to what we need to process tomorrow. We cannot just ``annotate our way out"~\cite{eisenstein:2013:bad}.  Moreover, it might not be trivial to find the right annotators; annotation schemes might need adaptation as well~\cite{zinsmeister:ea:2014} and tradeoffs for doing so need to be defined~\cite{schneider:2015}. 

What we need is \textit{quick ways to semi-automatically gather annotated data}, and use more unsupervised and weakly supervised approaches.

\subsection{Bring training and test data closer to each other}
The second  approach is based on the idea of making data resemble each other more. The first strategy here is \textit{normalization}, that is,  preprocess the input to make it closer to what our technology expects, e.g.~\newcite{han:baldwin:2013}. A less known but similar approach is to artificially corrupt the training data to make it more similar to the expected target domain~\cite{vanderplas:ea:2009}. However, normalization implies ``norm'', and as \newcite{eisenstein:2013:bad} remarks: whose norm are we targeting? (e.g., \textit{labor} vs \textit{labour}). Furthermore, he notes that it is surprisingly difficult to find a precise notion of the normalization task. 

Corrupting the training data is a less explored endeavor. This second strategy though hinges on the assumption that one knows what to expect. 

What we need are models that do provide nonsensical predictions on unexpected inputs, i.e., models that include \textit{invariant representations}. For example,  our models should be capable of learning similar representations for the same inherent concept, e.g., \textit{kiss} vs \textit{:*} or \textit{love} vs \textit{$<$3}. Recent shifts towards using sub-token level information can be seen as one step in this direction.

\subsection{Domain adaptation}
There is a large body of work on adapting models trained on some source domain to work better on some new target domain. Approaches range from feature augmentation, shared representation learning, instance weighting, to approaches that exploit representation induced from general background corpora. For an overview, see~\cite{plank:2011,weiss2016survey}. However, what all of these approaches have in common is an unrealistic assumption: \textit{they know the target domain}. That is, researchers typically have a small amount of target data available that they can use to adapt their models. 

An extreme case of adaptation is cross-lingual learning, whose goal is similar: adapt models trained on some source languages to languages in which few or no resources exist. Also here a large body of work assumes knowledge of the target language and requires some in-domain, typically parallel data. However, most work has focused on a restricted set of languages, only recently approaches emerged that aim to transfer from multiple sources to many target languages~\cite{agic:ea:2016}.

What we need are methods that can adapt quickly to unknown domains and languages, without much assumptions on what to expect, and use multiple sources, rather than just one. In addition, our models need to \textit{detect} when to trigger domain adaptation approaches.

%This bold opinion piece is not intended at a critique on our community, rather, as a reflection on assumptions made in the field, including my own previous work. 
In the next parts I will outline some possibilities to address these challenges. However, there are other important areas that I will not touch upon here (e.g., evaluation).

\section{Fortuitous data}

What we need are models that are more robust, work better on unexpected input and can be trained from semi-automatically or weakly annotated data, from a variety of sources.
In order to build such models, I argue that the key is to look for \textit{signal} in non-obvious places, i.e., \textit{fortuitous data}.\footnote{Thanks to Anders Johannsen for suggesting \textit{fortuitous} when I was in search for a name for \textit{serendipitous} \textit{casual} data.} 

Fortuitous data is data out there \textit{that just waits to be harvested}.  It might be in plain sight, but is neglected (available but not used), or it is in raw form and first needs to be refined (almost ready but needs refinement). \textit{Availability} and \textit{ease-of-use} (or \textit{readiness}) are therefore two important dimensions that define fortuitous data. Fortuitous data is the unintended yield of a process, a promising by-product or \textit{side benefit}.

\begin{table}
\begin{tabular}{lcc}
\toprule
Side benefit of: & availability & readiness \\
\midrule
User-generated content & + & + \\
Annotation & - & +\\
Behavior & + & - \\
\bottomrule
\end{tabular}
\caption{Typology of fortuitous data.}
\label{fig:typology}
\end{table}
In the following I will outline potential sources of fortuitous data. An overview is given in Table~\ref{fig:typology}.

\paragraph{Side benefit of user-generated content} This is data of high availability and high readiness, but it is often not used or ``preprocessed away''.
This source of fortuitous data includes user-generated content like webpages, social media posts, community-efforts like Wikipedia or Wiktionary. Concrete examples include hyperlinks that can be used to  build more robust named entity and part-of-speech taggers~\cite{plank:ea:14:coling}, or HTML markup for parsing~\cite{spitkovsky:ea:2010}. Similarly, Wiktionary can be used to mine large pools of data for unambiguous instances~\cite{hovy2015mining}, or can guide constrained inference like in type-constrained POS tagging~\cite{tackstrom:ea:2013,plank:ea:2014}. Broadly speaking, exploiting the web to process the web.
%Readiness: +, availability +

\paragraph{Side benefit of annotation} Another yield that is often disregarded is annotator disagreement. 
Such data has high readiness, but low availability. It is still rare for annotation efforts to release intermediate or preliminary stages of the annotation project, but such data contains precious signal. 

In fact, instead of adjudicating annotator decisions, we should embrace it. Annotator disagreement contains actual signal informative for a variety of tasks, including tagging, parsing,  supersense tagging and relation extraction, e.g.,~\cite{plank:ea:2014,aroyo2015truth}. % XXX (+Welty work). %Readiness: +, availability -

\paragraph{Side benefit of behavior} When people produce or read texts, they produce loads of by-product in form of behavior data. Examples here include click-through data, but also more distant sources such as cognitive processing data like eye tracking or keystroke dynamics. In a pilot study, I found  keystroke logs carry signal that can be used to inform NLP. Such data represents a potentially immense resource (imagine logging devices build into online editors or mobile phones, or eye tracking build into mobile devices). However, only very little work explored this source yet, e.g.,~\cite{barrett:ea:2015,klerke:ea:2016}. It is also the ``most distant'' fortuitous source, having high availability and low readiness, as data often first needs to be \textit{refined}.

\vspace*{0.2cm}
Using fortuitous data can thus be seen as a way to quickly obtain semi-automatically labeled data, from a variety of sources. 
If we pair fortuitous data with appropriate learning algorithms (transfer/multi-task learning), this will enable language technology that can adapt quickly to new language varieties. However, one question remains.

\section{But what's in a domain?}

As already noted earlier~\cite{plank:2011}, there is  \textit{no common ground on what constitutes a  domain}. 
\newcite{blitzer2006domain} attribute domain differences mostly to differences in vocabulary, \newcite{biber} explores differences between corpora from a sociolinguistics perspective. \newcite{mcclosky:2010} considers it in a broader view: ``By domain, we mean the style, genre, and medium of a document.'' Terms such as genre, register, text type, domain, style are often used differently in different communities~\cite{lee2002genres}, or interchangeably. 

While there exists no definition of domain, work on domain adaptation is plentiful but mostly focused on assuming a dichotomy: source versus target, without much interest in \textit{how} they differ. In fact, there is surprisingly little work on how texts \textit{vary} and the consequence for NLP. It is established that out-of-vocabulary (OOV) tokens impact NLP performance. However, what are other factors? 

Interest in this question re-emerged recently. For example, 
focusing on annotation difficulty, ~\newcite{zeldes-simonson:2016} remark ``that domain adaptation may be folding in \textit{sentence type} effects'', motivated by earlier findings by~\newcite{silveira2014gold} who remark that ``[t]he most striking difference between the two types of data [Web and newswire] has to do with imperatives, which occur two orders of magnitude more often in the EWT [English Web Treebank].'' A very recent paper examines word order properties and their impact on parsing taking a control experiment approach~\cite{gulordava2016multi}. On another angle, it has been shown that tagging accuracy correlates with demographic factors such as age~\cite{hovy:soegaard:2015:age}.

I want to propose that `domain' is an overloaded term. Besides the mathematical definition, in NLP it is typically used to refer to some coherent data with respect to topic or genre. However, there are many other (including yet unknown factors) out there, such as demographic factors, communicational purpose, but also sentence type, style, medium, technology/medium, language, etc. At the same time, these categories are not sharply defined either. Rather than imposing hard categories, let us consider a Wittgensteinian view.

\section{The variety space}
\label{sec:variationspace}
 I here propose to see a domain as \textit{variety} in a high-dimensional \textit{variety space}. Points in the space are the data instances, and regions form domains. A dataset $\mathcal{D}$ is a sample from the \textit{variety space}, conditioned on latent factors $V$:
 
 $$\mathcal{D} \sim P(X,Y|V)$$
 
\begin{figure}\centering
\includegraphics[width=0.6\columnwidth]{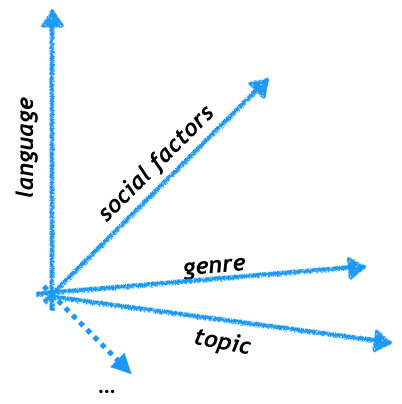}
\caption{What's in a \textit{domain}? Domain is an overloaded term. I propose to use the term \textit{variety}. A dataset is a sample from the \textit{variety space}, a unknown high-dimensional space, whose dimensions contain (fuzzy) aspects such as language (or dialect), topic or genre, and social factors (age, gender, personality, etc.), amongst others. A domain forms a region in this space, with some members more prototypical than others.}
\label{fig:variety}
\end{figure}

The variety space is a unknown high-dimensional space, whose dimensions (latent factors $V$) include (fuzzy) aspects such as language (or dialect), topic or genre, and social factors (age, gender, personality, etc.), amongst others.  A domain is a \textit{variety} that forms a region in this complicated network of similarities, with some members more prototypical than others. However, we have neither access to the number of latent factors nor to their types. This vision is inspired by the notion of prototype theory in Cognitive Science and Wittgenstein's \textit{graded notion} of categories. Figure~\ref{fig:variety} shows a hypothetical example of this variety space. 

Our datasets are subspaces of this high-dimensional space. Depending on our task, instances are sentences, documents etc. In the following I will use POS tagging as a running example to analyze what's in a domain, by referring to the datasets with the typically used categories.

\paragraph{Some empirical evidence - Taggers and Data} Let us examine two POS taggers representative for different tagging approaches and evaluate them on several varieties. We use  \textsc{TnT},\footnote{\url{http://www.coli.uni-saarland.de/~thorsten/tnt/}} an HMM-based tagger, and \textsc{Bilty}, a bidirectional LSTM tagger~\cite{plank:ea:2016}. Both taggers are trained on the WSJ training portion converted to Universal POS tags~\cite{petrov:ea:2012}. As test sets we consider parts of the Web Treebank (emails and answers), two Twitter datasets (\textsc{Foster} and \textsc{Gimpel/oct27}, Twitter sample 1 and 2 respectively), review data from two different age groups~\cite{hovy:soegaard:2015:age}, above 45 and below 35 years, and data from the CoNLL-X dataset from other Indogermanic languages.\footnote{\url{http://ilk.uvt.nl/conll/free_data.html} except Dutch because of joined MWU units.} These datasets were chosen to represent different varieties. 

\begin{table}
\resizebox{\columnwidth}{!}{
\begin{tabular}{ll|lll|c}
\toprule
\textsc{Variety} & \textsc{Sample}   & \textsc{TnT} & \textsc{Bilty}$_{\vec{w}}$ & \textsc{Bilty}$_{\vec{w}+\vec{c}}$ & \textsc{OOV} \\
\midrule
(in-dom.)         & wsj.test &  96.63 & 97.25 & 97.85 & 20\\
\midrule
\multirow{4}{*}{domain}  & anwers & 90.08 & 91.24 & 91.93 & 27 \\
           & emails & 91.03 & 89.81 &  92.20 & 29 \\
            & Tw (foster) & 90.25 & 92.47 & 92.26 & 28\\
           & Tw (oct27) & 65.98 & 66.37 & 67.16 & 52\\
\midrule
\multirow{2}{*}{age} & U35 & 86.11 & 85.06& 86.53 & 20\\
      & O45 & 86.73 & 85.81 & 87.70 & 22\\
\midrule

\multirow{3}{*}{language} & da & 35.25 & 37.85 & 38.00 & 89\\
         & pt & 24.99 & 43.50 & 47.33 & 93 \\  
         & sv & 33.13 & 39.80 & 37.09 & 92\\
\bottomrule
\end{tabular}
}
\caption{Tagging accuracy on various test set varieties (domains, languages and age groups; Tw=Twitter), using coarse POS~\cite{petrov:ea:2012}. OOV: out-of-vocabulary rate wrt \textsc{Wsj.train}. Accuracy is significantly correlated with OOV rate ($\rho$ = -0.70).}
\label{tbl:results}
\end{table}

\begin{figure}
\includegraphics[width=\columnwidth]{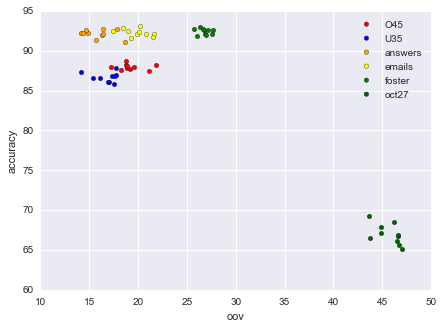}
\includegraphics[width=\columnwidth]{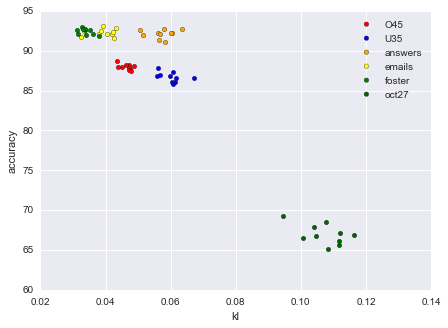}
\caption{Accuracy of the \textsc{Wsj} tagger on 10 bootstrap samples ($k=150$). Above: Accuracy versus OOV rate, Below: Accuracy vs KL divergence (src and trg gold POS bigram distributions). Different Twitter samples (green and darkgreen) exhibit very different behavior; oct27 has many OOVs \textit{and} a high KL div; \textsc{Foster} is much closer to WSJ in terms of KL div.}
\label{fig:acc}
\end{figure}

\paragraph{Results} Table~\ref{tbl:results} shows POS tagging accuracies. First, as is well known, we see that all taggers suffer when applied to other domains. However, models trained on \textsc{WSJ} fare worse on data from the younger age group, thus age is a covariate.  This confirms the age bias reported in~\cite{hovy:soegaard:2015:age} for the same data but using different taggers. If we stretch the notion of variety to other languages, we see that performance unsurprisingly drops dramatically. Remember, we just apply an in-domain single language tagger to other languages, although only trained on \textsc{WSJ} here.\footnote{Subtoken representations are used train a single tagger for multiple languages~\cite{gillick:ea:2015}.} \textsc{Bilty}$_{\vec{w}+\vec{c}}$ performs much better on other languages than \textsc{TnT}. Although the neural network-based tagger that uses both word and character embeddings fares better overall, both taggers suffer similarly, their accuracy variation is highly correlated ($\rho$=0.95, $p<0.01$ over all test sets; $\rho$=0.96 if we exclude the other languages, and $\rho$=0.94 if we also include \textsc{oct27}).

While the two age samples have similar OOV rates, the two Twitter samples differ substantially. Twitter sample 1 (\textsc{Foster}) has an OOV rate close to others (28), while sample 2 has the highest OOV (52), every other token is an OOV word. Thus, although both come from the same medium (Twitter), they are very different samples. In general,  OOV words are a major cause of performance drop. If we correlate all accuracies with OOV rate, we see a significant correlation ($\rho$ = -0.70, $p$=0.02274). However, caution is needed here, the high correlation could be influenced by outliers. In fact, if we exclude the other Twitter sample (\textsc{oct27}, which seems to form an outlier) and other languages, there is no significant correlation ($\rho$=0.23, $p$-val 0.6584), see Figure~\ref{fig:acc}, explained next.

Rather than just inspecting numbers of single test sets, we will now plot data characteristics versus accuracy. In order to do so 
we take 10 bootstrap samples ($k=150$ sentences) from the original test data, tag it with the best variant of \textsc{Bilty},  which uses word and character features, and evaluate it against gold POS. Figure~\ref{fig:acc} shows accuracy rates versus OOV rate (above) and accuracy vs KL-divergence between gold and predicted tag bigram distributions (lower plot). Each data point in the plot is a bootstrap sample. 

The plots show that Twitter sample 2 (dark green, \textsc{Foster}) is similar in OOV rate to emails and answers; In fact, it is very close to the original dataset (\textsc{WSJ}), it differs the \textit{least} from \textsc{WSJ} in terms of POS KL-divergence (lower plot). In contrast, Twitter sample 2 (green, \textsc{oct27}) has not only high OOV rate, but it also differs highly in KL div from \textsc{WSJ}.  The dataset contains many unusual POS sequences that are hard to predict. The same is true for age, the KL plot confirms that the tags of the younger group are harder to predict.

We see that performance varies greatly on different samples of Twitter data, as also reported earlier~\cite{hovy2014when}. This suggest that 
Twitter is not a `single domain'. It spans an entire range of varieties (social groups, agents, topics, even languages, etc.).  Relating back to variety space, it seems that our two samples span different subspaces. Although the two samples used here do not resemble each other, they still share the commonality of being drawn from the same category (in this case, medium), mirroring Wittgenstein's theory of family resemblance, cf.~\cite{givon1986prototypes}. In fact, if we think about data from Twitter, we will have a prototypical member in mind, but members might vary highly. Whenever we build models for, say, Twitter, we need to be aware of these properties. The more the data varies, the more test samples we will need to achieve higher confidence in our models.

\section{Conclusions}

Current NLP models still suffer dramatically when applied to non-canonical data, where canonicity 
is a relative notion; in our field, newswire was and still often is the de-facto standard, the canonical data we typically train our models on.

While newswire has advanced the field in so many ways, it has also introduced almost imperceptible biases. What we need is to be aware of such biases, collect enough biased data, and model \textit{variety}. I argue that if we embrace the variety of this heterogeneous data by combining it with proper algorithms, in addition to including text covariates/latent factors, we will not only produce more robust models, but will also enable adaptive language technology capable of addressing natural language variation.

\section*{Acknowledgments}

I would like to thank the organizers for the invitation
to the keynote at KONVENS 2016. I am also grateful to H\'{e}ctor Mart\'{i}nez Alonso, Dirk Hovy, Anders Johannsen, Zeljko Agi\'{c} and Gertjan van Noord for valuable discussions and feedback on earlier drafts of this paper.

\bibliographystyle{konvens2016}
\bibliography{biblio.bib}

\end{document}